\documentclass[11pt,letterpaper]{article}

\usepackage{scrextend}

\usepackage[auth-lg]{authblk}
\usepackage{emnlp2017}
\usepackage{times}
\usepackage{latexsym}
\usepackage{url}
\usepackage{amssymb}
\usepackage{amsmath}
\usepackage{dsfont}
\usepackage{graphicx}
\usepackage{multirow}
\usepackage{tikz,pgfplots}
\usepackage{balance}

\emnlpfinalcopy

\def\emnlppaperid{***}

\usepackage{amsthm}

\newtheorem*{definition*}{Definition}
\newtheorem*{theorem*}{Theorem}

\usepackage{titlesec}
\titlespacing*{\section} {0pt}{1.2ex plus 1ex minus .2ex}{0.8ex plus .2ex}
\titlespacing*{\paragraph} {0pt}{0.2ex plus 0.05ex minus .05ex}{1em}

\usepackage{caption}
\DeclareCaptionFormat{capfont}{\fontsize{10}{6}\selectfont#1#2#3}
\usepackage{subcaption}
\makeatletter
\g@addto@macro\small{%
  \setlength\abovedisplayskip{-5pt}
  \setlength\abovedisplayshortskip{-5pt}
  \setlength\belowdisplayshortskip{-7pt}
  \setlength\belowdisplayskip{-7pt}
}
\makeatother

\date{}

\usepackage{algorithmicx} 
\usepackage[noend]{algpseudocode}
\usepackage{algorithm}
\usepackage[T1]{fontenc}

\algnewcommand\algorithmicdefinitions{\textbf{Definitions:}}
\algnewcommand\Definitions{\item[\algorithmicdefinitions]}
\renewcommand{\algorithmiccomment}[1]{{\color{gray}\raisebox{1px}{\texttt{\guillemotright}} #1}}
\algnewcommand{\LineComment}[1]{\Statex \hskip\ALG@thistlm \algorithmiccomment{#1}}
\algrenewcommand\alglinenumber[1]{\footnotesize #1:}
\algrenewcommand\algorithmicindent{1.0em}%
\makeatletter
\newcommand{\StatexIndent}[1][3]{%
  \setlength\@tempdima{\algorithmicindent}%
  \Statex\hskip\dimexpr#1\@tempdima\relax}

\usepackage{arydshln}
\newcommand{\dline}{\hdashline[0.5pt/1pt]}

\newcommand{\eat}[1]{}

\newcommand{\edit}[3]{\textcolor{#1}{#2}}

\newcommand{\jle}[2]{\edit{orange}{#1}{#2}}

\newcommand{\reals}{\mathds{R}}
\newcommand{\nlstring}[1]{{\em #1}}

\newcommand{\policy}{\pi}
\newcommand{\state}{s}
\newcommand{\allstate}{\mathcal{S}}
\newcommand{\startstate}{s_{1}}
\newcommand{\goalstate}{s_{g}}
\newcommand{\instruction}{\bar{x}}
\newcommand{\allinstruction}{\mathcal{X}}
\newcommand{\token}{x}
\newcommand{\action}{a}
\newcommand{\allaction}{\mathcal{A}}
\newcommand{\transfunc}{T}
\newcommand{\genimage}{\textsc{Img}}
\newcommand{\image}{I}

\newcommand{\act}[1]{{\tt \MakeUppercase{#1}}}
\newcommand{\stopaction}{\act{stop}}
\newcommand{\execution}{\bar{e}}

\newcommand{\exechorizon}{J}

\newcommand{\conv}{\textsc{CNN}}
\newcommand{\lstm}{\textsc{LSTM}}
\newcommand{\lstmrep}{\mathbf{l}}
\newcommand{\langrep}{\mathbf{\instruction}}
\newcommand{\mlplayer}[1]{\mathbf{h}^{#1}}
\newcommand{\statehorizon}{K}
\newcommand{\visualrep}{\mathbf{v}}
\newcommand{\ostate}{\tilde{s}}

\newcommand{\ostatevec}{\mathbf{\ostate}}

\newcommand{\param}{\theta}
\newcommand{\embed}{\psi}
\newcommand{\tokenembed}{\embed}
\newcommand{\actionembed}{\embed_a}

\newcommand{\reward}{R}
\newcommand{\epochs}{T}
\newcommand{\rewardpenalty}{\delta}
\newcommand{\entropycoef}{\lambda}
\newcommand{\objective}{\mathcal{J}}
\newcommand{\shaping}{F}

\title{Mapping Instructions and Visual Observations to Actions \\ with Reinforcement Learning}

\author[$\dagger$]{Dipendra Misra}
\author[$\ddagger$]{John Langford}
\author[$\dagger$]{Yoav Artzi}
\affil[$\dagger$]{
Dept. of Computer Science and Cornell Tech\\
Cornell University\\ 
New York, NY 10044 \authorcr
{\tt \{dkm, yoav\}@cs.cornell.edu}
\authorcr
}
\affil[$\ddagger$]{
Microsoft Research\\
New York, NY 10011 \authorcr
{\tt jcl@microsoft.com}
}

\begin{document}
\maketitle

\begin{abstract}

We propose to directly map raw visual observations and text input to actions for instruction execution. While existing approaches assume access to structured environment representations or use a pipeline of separately trained models, we learn a single model to jointly reason about linguistic and visual input. We use reinforcement learning in a contextual bandit setting to train a neural network agent. To guide the agent's exploration, we use reward shaping with different forms of supervision. Our approach does not require intermediate representations, planning procedures, or training different models. We evaluate in a simulated environment, and show significant improvements over supervised learning and common reinforcement learning variants.

\end{abstract}

\section{Introduction}
\label{sec:intro}

\begin{figure}
	\centering
	\includegraphics[width=0.95\linewidth,clip,trim=0 30 0 9]{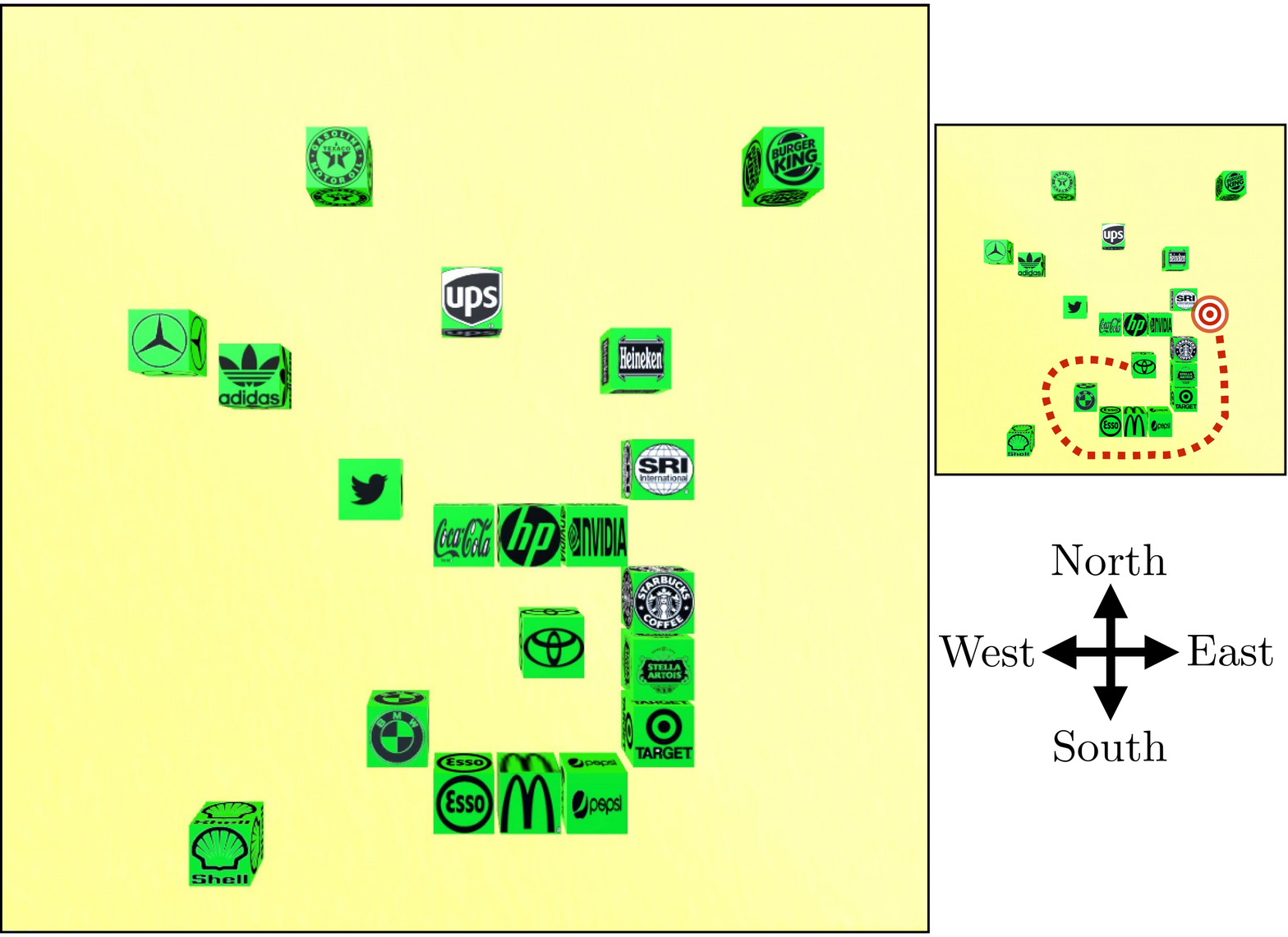}
	\vspace{-10pt}
	\begin{center}
	\footnotesize
	\begin{tabular}{|p{0.95\linewidth}|}
		\hline
		\nlstring{Put the Toyota block in the same row as the SRI block, in the first open space to the right of the SRI block} \\ \dline
		\nlstring{Move Toyota to the immediate right of SRI, evenly aligned and slightly separated} \\ \dline
		\dline
		\nlstring{Move the Toyota block around the pile and place it just to the right of the SRI block} \\ \dline
		\nlstring{Place Toyota block just to the right of The SRI Block} \\ 
		\dline
		\nlstring{Toyota, right side of SRI} \\ 
		\hline
	\end{tabular}
	\end{center}
	\caption{Instructions in the Blocks environment. The instructions all describe the same task. Given the observed RGB image of the start state (large image), our goal is to execute such instructions. In this task,  the direct-line path to the target position is blocked, and the agent must plan and move the Toyota block around. The small image marks the target and an example path, which includes 34 steps.}
\label{fig:instruction}
\vspace{-5pt}
\end{figure}

An agent executing natural language instructions requires robust understanding of language and its environment. 
Existing approaches addressing this problem  assume structured environment representations~\cite[e.g.,.][]{Chen:11,Mei:16neuralnavi}, or combine separately trained models~\cite[e.g.,][]{Matuszek:10,Tellex:11}, including for language understanding and visual reasoning. 
We propose to directly map text and raw image input to actions with a single learned model.
This approach offers multiple benefits, such as not requiring intermediate representations, planning procedures, or training multiple models.

Figure~\ref{fig:instruction} illustrates the problem in the Blocks environment~\cite{Bisk:16nl-robots}. 
The agent observes the environment as an RGB image using a camera sensor. 
Given the RGB input, the agent must recognize the blocks and their layout.
To understand the instruction, the agent must identify the block to move (Toyota block) and the destination (just right of the SRI block). 
This requires solving semantic and grounding problems. 
For example, consider the topmost instruction in the figure.
The agent needs to identify the phrase referring to the block to move, \nlstring{Toyota block}, and ground it.  
It must resolve and ground the phrase \nlstring{SRI block} as a reference position, which is then modified by the spatial meaning recovered from \nlstring{the same row as} or \nlstring{first open space to the right of}, to identify the goal position. 
Finally, the agent needs to generate actions, for example moving the Toyota block around obstructing blocks.

To address these challenges with a single model, we design a neural network agent. 
The agent executes instructions by generating a sequence of actions.  
At each step, the agent takes as input the instruction text, observes the world as an RGB image, and selects the next action. 
Action execution changes the state of the world. 
Given an observation of the new world state, the agent selects the next action. 
This process continues until the agent indicates execution completion. 
When selecting actions, the agent jointly reasons about its observations and the instruction text. 
This enables decisions based on close interaction between observations and linguistic input.

We train the agent with different levels of supervision, including complete demonstrations of the desired behavior and annotations of the goal state only. 
While the learning problem can be easily cast as a supervised learning problem, learning only from the states observed in the training data results in poor generalization and failure to recover from test errors. 
We use reinforcement learning~\cite{Sutton:98rl} to observe a broader set of states through exploration. 
Following recent work in robotics~\cite[e.g.,][]{Levine:16endtoend,Rusu:16sim-to-real}, we assume the training environment, in contrast to the test environment, is instrumented and provides access to the state. 
This enables a simple problem reward function that uses the state and provides positive reward on task completion only. 
This type of reward offers two important advantages: (a) it is a simple way to express the ideal agent behavior we wish to achieve, and (b) it creates a platform to add training data information. 

We use reward shaping~\cite{Ng:99rewardshaping} to exploit the training data and add to the reward additional information. 
The modularity of shaping allows varying the amount of supervision, for example by using complete demonstrations for only a fraction of the training examples. 
Shaping also naturally associates actions with immediate reward. 
This enables learning in a contextual bandit setting~\cite{AuerCFS02,Langford:07}, where optimizing the immediate reward is sufficient and has better sample complexity than unconstrained reinforcement learning~\cite{Agarwal:14taming-monster}.

We evaluate with the block world environment and data of \citet{Bisk:16nl-robots}, where each instruction moves one block (Figure~\ref{fig:instruction}). 
While the original task focused on source and target prediction only, we build an interactive simulator and formulate the task of predicting the complete sequence of actions.
At each step, the agent must select between 81 actions with 15.4 steps required to complete a task on average, significantly more than existing environments~\cite[e.g.,][]{Chen:11}.
Our experiments demonstrate that our reinforcement learning approach effectively reduces execution error by 24\% over standard supervised learning and 34-39\% over common reinforcement learning techniques. 
Our simulator, code, models,  and execution videos are available at: \url{https://github.com/clic-lab/blocks}.

\section{Technical Overview}

\paragraph{Task} 

Let $\allinstruction$ be the set of all \emph{instructions}, $\allstate$ the set of all \emph{world states}, and $\allaction$ the set of all \emph{actions}.  
An instruction $\instruction \in \allinstruction$ is a sequence $\langle \token_1, \dots, \token_n \rangle$, where each $\token_i$ is a token. 
The agent executes instructions by generating a sequence of actions, and indicates execution completion with the special action $\stopaction$.
Action execution modifies the world state following a transition function $\transfunc : \allstate \times \allaction \rightarrow \allstate$. 
The execution $\execution$ of an instruction $\instruction$ starting from $\startstate$ is an $m$-length sequence $\langle (\state_1, \action_1), \dots, (\state_m, \action_m)  \rangle$, where $\state_j \in \allstate$, $\action_j \in \allaction$, $\transfunc(\state_j, \action_j) = \state_{j+1}$ and $\action_m = \stopaction$. 
In Blocks (Figure~\ref{fig:instruction}), a state specifies the positions of all blocks. 
For each action, the agent moves a single block on the plane in one of four directions (north, south, east, or west). 
There are $20$ blocks, and $81$ possible actions at each step, including  $\stopaction$. 
For example, to correctly  execute the instructions in the figure, the agent's likely first action is $\act{TOYOTA\text{-}WEST}$, which moves the Toyota block one step west. 
Blocks can not move over or through other blocks.

\paragraph{Model}

The agent observes the world state via a visual sensor (i.e., a camera). 
Given a world state $\state$, the agent observes an RGB image $\image$ generated by the function $\genimage(\state)$. 
We distinguish between the world state $\state$ and the \emph{agent context}\footnote{We use the term \emph{context} similar to how it is used in the contextual bandit literature to refer to the information available for decision making. While agent contexts capture information about the world state, they do not include physical information, except as captured by observed images.} $\ostate$, which includes the instruction, the observed image $\genimage(\state)$, images of previous states, and the previous action. 
To map instructions to actions, the agent reasons about the agent context $\ostate$ to generate a sequence of actions. 
At each step, the agent generates a single action. 
We model the agent with a neural network policy. 
At each step $j$, the network takes as input the current agent context $\ostate_j$, and predicts the next action to execute $\action_j$. 
We formally define the agent context and model in Section~\ref{sec:model}. 

\paragraph{Learning} We assume access to training data with $N$ examples $\{ (\instruction^{(i)}, \startstate^{(i)}, \execution^{(i)})\}_{i=1}^N$, where $\instruction^{(i)}$ is an instruction, $\startstate^{(i)}$ is a start state, and $\execution^{(i)}$ is an execution demonstration of $\instruction^{(i)}$ starting at $\startstate^{(i)}$.  
We use policy gradient (Section~\ref{sec:learn}) with reward shaping derived from the training data to increase learning speed and exploration effectiveness (Section~\ref{sec:reshape}). 
Following work in robotics~\cite[e.g.,][]{Levine:16endtoend}, we assume an instrumented environment with access to the world state to compute the reward during training only. 
We define our approach in general terms with demonstrations, but also experiment with training using  goal states.

\paragraph{Evaluation} 
We evaluate task completion error on a test set $\{(\instruction^{(i)}, \startstate^{(i)}, \goalstate^{(i)}) \}^M_{i=1}$, where $\instruction^{(i)}$ is an instruction, $\startstate^{(i)}$ is a start state, and $\goalstate^{(i)}$ is the goal state. 
We measure execution error as the distance between the final execution state and $\goalstate^{(i)}$.

\section{Related Work}
\label{sec:related}

Learning to follow instructions was studied extensively with structured environment representations, including with semantic parsing~\cite{Chen:11,Kim:12,Kim:13,Artzi:13,Artzi:14,Artzi:14programmingdemo,Misra:15highlevel,Misra:16telldave}, alignment models~\cite{Andreas:15navi}, reinforcement learning~\cite{Branavan:09,Branavan:10,Vogel:10}, and neural network models~\cite{Mei:16neuralnavi}. 
In contrast, we study the problem of an agent that takes as input instructions and raw visual input. 
Instruction following with visual input was studied with pipeline approaches that use separately learned models for visual reasoning~\cite{Matuszek:10,Matuszek:12b,Tellex:11,Paul:16grounding}.
Rather than decomposing the problem, we adopt a single-model approach and learn from instructions paired with demonstrations or goal states. 
Our work is related to \citet{Sung:15robobarista}. While they use sensory input to select and adjust a trajectory observed during training, we are not restricted to training sequences. 
Executing instructions in non-learning settings has also received significant attention~\citep[e.g.,][]{Winograd:72,Webber:95,MacMahon:06}. 

Our work is related to a growing interest in problems that combine language and vision, including visual question answering~\cite[e.g.,][]{Antol:15vqa,Andreas:16nmn,Andreas:16compose}, caption generation~\cite[e.g.,][]{Chen:15coco,Chen:16coco,Xu:15visattn}, and visual reasoning~\cite{Johnson:16clevr,Suhr:17visual-reason}.
We address the prediction of the next action  given a world image and an instruction. 

Reinforcement learning with neural networks has been used for various NLP tasks, including text-based games~\cite{Narasimhan:15text-games,He:16deep-rl-language}, information extraction~\cite{Narasimhan:16rl-ie}, co-reference resolution~\cite{Clark:16drl-coref}, and dialog~\cite{Li:16drlchatbot}.

Neural network reinforcement learning techniques have been recently studied  for behavior learning tasks, including playing games~\cite{Mnih:13atari,Mnih:15humanlevelatari,Mnih:16async-deep-rl,Silver:16mastering-go} and solving memory puzzles~\cite{Oh:16rl-minecraft}.
In contrast to this line of work, our data is limited. 
Observing new states in a computer game simply requires playing it.
However, our agent also considers  natural language instructions. 
As the set of instructions is limited to the training data, the set of agent contexts seen during learning is constrained. 
We address the data efficiency problem by learning in a contextual bandit setting, which is known to be more tractable~\cite{Agarwal:14taming-monster}, and using reward shaping to increase exploration effectiveness. 
\citet{Zhu:16TargetdrivenVN} address generalization of reinforcement learning to new target goals in visual search by providing the agent an image of the goal state.
We address a related problem. However, we provide natural language and the agent must learn to recognize the goal state. 

Reinforcement learning is extensively used in robotics~\cite{Kober:13robotics-rl}.
Similar to recent work on learning neural network policies for robot control~\cite{Levine:16endtoend,Schulman:15trpo,Rusu:16sim-to-real}, we assume an instrumented training environment and use the state to compute rewards during learning.  
Our approach adds the ability to specify tasks using natural language.

\section{Model}
\label{sec:model}

\begin{figure*}
\centering
\includegraphics[width=0.8\textwidth,clip,trim=0 452 231 18]{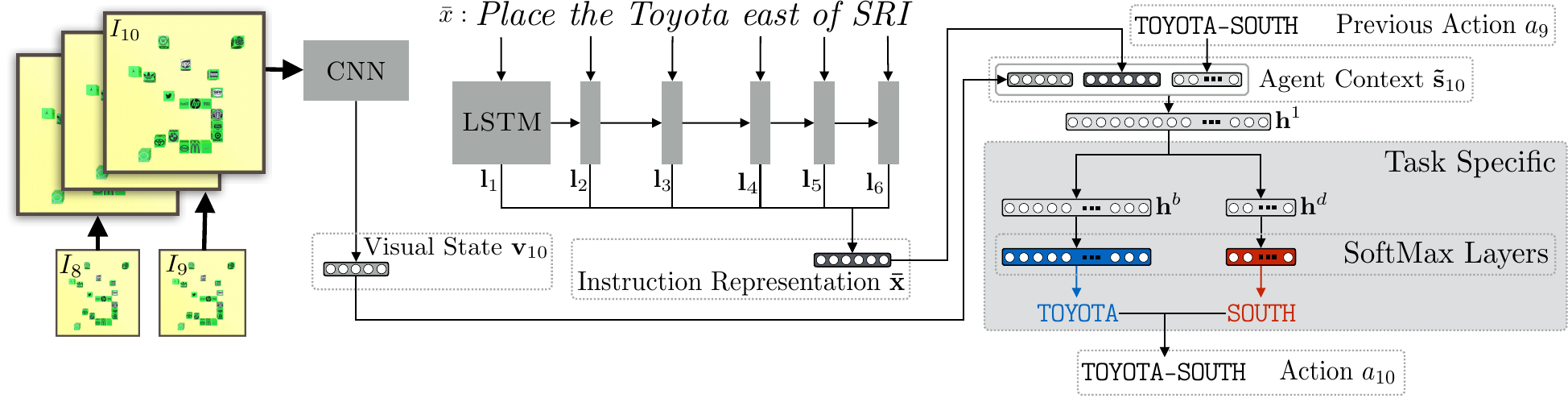}
\caption{Illustration of the policy architecture showing the 10th step in the execution of the instruction \nlstring{Place the Toyota east of SRI} in the state from Figure~\ref{fig:instruction}.  The network takes as input the instruction $\instruction$, image of the current state $\image_{10}$, images of previous states $\image_8$ and $\image_9$ (with $\statehorizon = 2$), and the previous action $\action_9$. The text and images are embedded with $\lstm$ and $\conv$. The actions are selected with the task specific multi-layer perceptron.}
\label{fig:arch}
\end{figure*}

We model the agent policy $\policy$ with a neural network. 
The agent observes the instruction and an RGB image of the world. 
Given a world state $\state$, the image $\image$ is generated using the function $\genimage(\state)$. 
The instruction execution is generated one step at a time. 
At each step $j$, the agent observes an image $\image_j$ of the current world state $\state_j$ and the instruction $\instruction$, predicts the action $\action_j$, and executes it to transition to the next state $\state_{j+1}$. 
This process continues until $\stopaction$ is predicted and the agent stops, indicating instruction completion. 
The agent also has access to $\statehorizon$ images of previous states and the previous action to  distinguish between different stages of the execution~\cite{Mnih:15humanlevelatari}. 
Figure~\ref{fig:arch} illustrates our architecture.

Formally,\footnote{We use bold-face capital letters for matrices and bold-face lowercase letters for vectors. Computed input and state representations use bold versions of the symbols. For example, $\langrep$ is the computed representation of an instruction  $\instruction$.} at step $j$, the agent considers an agent context $\ostate_j$, which is a tuple $(\instruction, \image_j, \image_{j-1},\dots,\image_{j-\statehorizon}, \action_{j-1})$, where $\instruction$ is the natural language instruction, $\image_j$ is an image of the current world state, the images $\image_{j-1},\dots,\image_{j-\statehorizon}$ represent $\statehorizon$ previous states, and $\action_{j-1}$ is the previous action. 
The agent context includes information about the current state and the execution. 
Considering the previous action $\action_{j-1}$ allows the agent to avoid repeating failed actions, for example when trying to move in the direction of an obstacle. 
In Figure~\ref{fig:arch}, the agent is given the instruction \nlstring{Place the Toyota east of SRI}, is  at the $10$-th execution step, and considers $\statehorizon=2$ previous images. 

We generate continuous vector representations for all inputs, and jointly reason about both text and image modalities to select the next action. 
We use a recurrent neural network~\cite[RNN;][]{Elman:90rnn} with a long short-term memory~\cite[LSTM;][]{Hochreiter:97lstm} recurrence to map the instruction $\instruction = \langle \token_1, \dots, \token_n\rangle$ to a vector representation $\langrep$.
Each token $\token_i$ is mapped to a fixed dimensional vector with the learned embedding function $\tokenembed(\token_i)$. 
The instruction representation $\langrep$ is computed by applying the $\lstm$ recurrence to generate a sequence of hidden states \mbox{$\lstmrep_i = \lstm(\tokenembed(\token_i), \lstmrep_{i-1})$}, and computing the mean \mbox{$\langrep = \frac{1}{n}\sum_{i=1}^{n} \lstmrep_i$}~\cite{Narasimhan:15text-games}. 
The current image $\image_j$ and previous images $\image_{j-1}$,\dots,$\image_{j-\statehorizon}$ are concatenated %
along the channel dimension and embedded with a convolutional neural network ($\conv$) to generate the visual state $\visualrep$~\cite{Mnih:13atari}. 
The last action $\action_{j-1}$ is embedded with the  function $\actionembed(\action_{j-1})$. 
The vectors $\visualrep_j$, $\langrep$, and $\actionembed(\action_{j-1})$ are concatenated to create the agent context vector representation \mbox{$\ostatevec_j  = [\visualrep_j, \langrep, \actionembed(\action_{j-1})]$}.

To compute the action to execute, we use a feed-forward perceptron that decomposes according to the domain actions. 
This computation selects the next action  conditioned on the instruction text and observations from both the current world state and recent history. 
In the block world domain, where actions decompose to selecting the block to move and the direction, the network  computes block and direction probabilities. 
Formally, we decompose an action $\action$ to direction $\action^D$ and block $\action^B$. 
We compute the feedforward network:

\begin{small}
\begin{eqnarray}
\nonumber	\mlplayer{1} &=& \max(\mathbf{W}^{(1)}\ostatevec_j + \mathbf{b}^{(1)}, 0) \\
\nonumber	\mlplayer{D} &=& \mathbf{W}^{(D)}\mlplayer{1} + \mathbf{b}^{(D)} \\
\nonumber	\mlplayer{B} &=& \mathbf{W}^{(B)}\mlplayer{1}+ \mathbf{b}^{(B)}\;\;,
\end{eqnarray}	
\end{small}

\noindent
and the action probability is a product of the component probabilities:

\begin{small}
\begin{eqnarray}
\nonumber	P(\action_j^D = d \mid \instruction, \state_j, \action_{j-1}) &\propto & \exp(\mlplayer{D}_{d}) \\
\nonumber	P(\action_j^B = b \mid \instruction, \state_j,  \action_{j-1}) &\propto & \exp(\mlplayer{B}_{b})\;\;.
\end{eqnarray}	
\end{small}

At the beginning of execution, the first action $\action_0$ is set to the special value {\tt NONE}, and previous images are zero matrices.  
The embedding function $\tokenembed$ is a learned matrix. 
The function $\actionembed$ concatenates the embeddings of $\action^D_{j-1}$ and $\action^B_{j-1}$, which are obtained from learned matrices, to compute the embedding of $\action_{j-1}$. 
The model parameters $\param$ include $\mathbf{W}^{(1)}$, $\mathbf{b}^{(1)}$, $\mathbf{W}^{(D)}$, $\mathbf{b}^{(D)}$, $\mathbf{W}^{(B)}$, $\mathbf{b}^{(B)}$, the parameters of the $\lstm$ recurrence, the parameters of the convolutional network $\conv$, and the embedding matrices. In our experiments (Section~\ref{sec:exp}), all parameters are learned without external resources.

\section{Learning}
\label{sec:learn}

We use policy gradient for reinforcement learning~\cite{Williams:92reinforce} to estimate the parameters $\param$ of the agent policy. 
We assume access to a training set of $N$ examples $\{ (\instruction^{(i)}, \startstate^{(i)}, \execution^{(i)})\}_{i=1}^N$, where $\instruction^{(i)}$ is an instruction, $\startstate^{(i)}$ is a start state, and $\execution^{(i)}$ is an execution demonstration starting from $\startstate^{(i)}$ of instruction $\instruction^{(i)}$. 
The main learning challenge is learning how to execute instructions given raw visual input from relatively limited data. 
We learn in a contextual bandit setting, which provides theoretical advantages over general reinforcement learning. 
In Section~\ref{sec:results}, we verify this empirically. 

\paragraph{Reward Function} 

The instruction execution problem defines a simple problem reward to measure task completion.  
The agent receives a positive reward when the task is completed, a negative reward for incorrect completion (i.e., $\stopaction$ in the wrong state) and actions that fail to execute (e.g., when the direction is blocked), and a small penalty otherwise,  which induces a preference for shorter trajectories. 
To compute the reward, we assume access to the world state. 
This learning setup is inspired by work in robotics, where it is achieved by instrumenting the training environment (Section~\ref{sec:related}). 
The agent, on the other hand, only uses the agent context (Section~\ref{sec:model}). 
When deployed, the system relies on visual observations and natural language instructions only. 
The reward function \mbox{$\reward^{(i)} : \allstate \times \allaction \rightarrow
\reals$} is defined for each training example $(\instruction^{(i)}, \startstate^{(i)}, \execution^{(i)})$, $i = 1\dots N$:

\begin{small}
\begin{equation}
\nonumber	\reward^{(i)}(\state, \action) = \begin{cases} 1.0 & \text{if } \state = \state_{m^{(i)}} \text{ and } \action = \stopaction \\	 -1.0 & \state \neq \state_{m^{(i)}} \text{ and } \action = \stopaction \\ -1.0 & \action \text{ fails to execute} \\ -\rewardpenalty & \text{else} \end{cases}\;\;,
\end{equation}
\end{small}

\noindent
where $m^{(i)}$ is the length of $\execution^{(i)}$. 

The reward function does not provide intermediate positive feedback to the agent for actions that bring it closer to its goal. 
When the agent explores randomly early during learning, it is unlikely to encounter the goal state due to the large number of steps required to execute tasks. 
As a result, the agent does not observe positive reward and fails to learn. 
In Section~\ref{sec:reshape}, we describe how reward shaping, a method to augment the reward with additional information, is used to take advantage of the training data and address this challenge. 

\paragraph{Policy Gradient Objective}

We adapt the policy gradient objective defined by \citet{Sutton:99policygradint} to multiple starting states and reward functions:

\begin{small}
\begin{equation}
\nonumber \objective = \frac{1}{N} \sum_{i=1}^N V^{(i)}_\policy(\startstate^{(i)} ) \;\;,
\end{equation}
\end{small}

\noindent
where $V^{(i)}_\policy(\startstate^{(i)})$ is the value given by $\reward^{(i)}$ starting from  $\startstate^{(i)}$ under the policy $\policy$. 
The summation expresses the goal of learning a behavior parameterized by natural language instructions.

\paragraph{Contextual Bandit Setting}

In contrast to most policy gradient approaches, we apply the objective to a contextual bandit setting where immediate reward is optimized rather than total expected reward. 
The primary theoretical advantage of contextual bandits is much tighter sample complexity bounds when comparing upper bounds for contextual bandits~\cite{Langford:07} even with an adversarial sequence of contexts~\cite{AuerCFS02} to lower bounds~\cite{Krishnamurthy:16PACRL} or upper bounds~\cite{Kearns:99sparse-sample-complexity} for total reward maximization.  
This property is particularly suitable for the few-sample regime common in natural language problems. 
While reinforcement learning with neural network policies is known to require large amounts of training data~\cite{Mnih:15humanlevelatari},  the limited number of training sentences constrains the diversity and volume of agent contexts we can observe during training. 
Empirically, this translates to poor results when optimizing the total reward ($\textsc{REINFORCE}$ baseline in Section~\ref{sec:results}).
To derive the approximate gradient, we use the likelihood ratio method:

\begin{small}
\begin{equation}
\nonumber \nabla_\theta \objective =  \frac{1}{N}\sum_{i=1}^N \mathbb{E}[ \nabla_\theta \log \policy(\ostate, \action) \reward^{(i)}(\state, \action)]\;\;,
\end{equation}
\end{small}

\noindent
where reward is computed from the world state but policy is learned on the agent context. We approximate the gradient using sampling. 

This training regime, where immediate reward optimization is sufficient to optimize policy parameters $\param$, is enabled by the shaped reward we introduce in Section~\ref{sec:reshape}.
While the objective is designed to work best with the shaped reward, the algorithm remains the same for any choice of reward definition including the original problem reward or several possibilities formed by reward shaping. 

\paragraph{Entropy Penalty} 

We observe that early in training, the agent is overwhelmed with negative reward and rarely completes the task. 
This results in the policy $\policy$ rapidly converging towards a suboptimal deterministic policy with an entropy of $0$. 
To delay premature convergence we add an entropy term to the objective~\cite{Williams:91function-opt-rl,Mnih:16async-deep-rl}.
The entropy term encourages a uniform distribution policy, and in practice stimulates exploration early during training. 
The regularized gradient is: 

\begin{small}
\begin{eqnarray}
\nonumber  \nabla_\theta  \objective =& \\ \nonumber & \hspace{-28pt}\displaystyle \frac{1}{N}\sum_{i=1}^N \mathbb{E}[ \nabla_\theta \log \policy(\ostate, \action) \reward^{(i)}(\state, \action) + \entropycoef\nabla_\theta  H(\policy(\ostate, \cdot))]\;\;,
\end{eqnarray}	
\end{small}

\noindent
where $H(\policy(\ostate, \cdot))$ is the entropy of $\policy$ given the agent context $\ostate$, $\entropycoef$ is a hyperparameter that controls the strength of the regularization. 
While the entropy term delays premature convergence, it does not eliminate it. 
Similar issues are observed for vanilla policy gradient~\cite{Mnih:16async-deep-rl}.

\begin{figure}[t!]
\begin{tabular}{@{}p{7cm}}
\vspace{-0.275in}
\begin{algorithm}[H]
\caption{Policy gradient learning}
\begin{algorithmic}[1]
\footnotesize
\Require Training set $\{(\instruction^{(i)}, \startstate^{(i)}, \execution^{(i)})\}_{i=1}^N$,  learning rate $\mu$, epochs $\epochs$, horizon $\exechorizon$, and entropy regularization term $\entropycoef$. %
\Definitions $\genimage(\state)$ is a camera sensor that reports an RGB image of state $\state$. $\policy$ is a probabilistic neural network policy parameterized by $\param$, as described in Section~\ref{sec:model}. $\textsc{Execute}(\state, \action)$ executes the action $\action$ at the state $\state$, and returns the new state. $\reward^{(i)}$ is the reward function for example $i$. $\textsc{Adam}(\Delta)$ applies a per-feature learning rate to the gradient $\Delta$~\citep{Kingma:14adam}.
\Ensure Policy parameters $\param$.
\State \Comment{Iterate over the training data.}
\For{$t=1$ to $\epochs$, $i=1$ to $N$}\label{alg:learn:dataloop:start}
\State $\image_{1-K},\dots,\image_0 = \vec{0}$
\State $\action_0 = {\tt NONE}$, $\state_1 = \startstate^{(i)}$
\State $j = 1$
\State \Comment{Rollout up to episode limit.}
\While{$j \leq \exechorizon \text{ and } \action_j \neq \stopaction$}\label{alg:learn:rollout:start}
\State \Comment{Observe world and construct agent context.}
\State $\image_j = \genimage(\state_j)$ \label{alg:learn:image}
\State $\ostate_j = (\instruction^{(i)},\image_j,  \image_{j-1},\dots,\image_{j-\statehorizon},\action_{j-1}^d )$\label{alg:learn:observe}
\State \Comment{Sample an action from the policy.}
\State $\action_j \sim \policy(\ostate_j, \action)$\label{alg:learn:sample}
\State $\state_{j+1} = \textsc{execute}(\state_j, \action_j)$\label{alg:learn:execute}
\State \Comment{Compute the approximate gradient.}
\State $\Delta_j \gets \nabla_\theta \log \policy(\ostate_j, \action_j) \reward^{(i)}(\state_j, \action_j)$\label{alg:learn:backpropagate}
\StatexIndent{\hspace{50pt} $+ \entropycoef\nabla_\theta  H(\policy(\ostate_j, \cdot))$}
\State $j += 1$
\EndWhile\label{alg:learn:rollout:end}
\State $\param \gets \param + \mu\textsc{Adam}(\frac{1}{j}\sum_{j'=1}^j\Delta_{j'})$\label{alg:learn:paramupdate}
\EndFor\label{alg:learn:dataloop:end}
\State \textbf{return} $\param$ \label{alg:main:return}
\end{algorithmic}
\label{alg:learn} 
\end{algorithm}
\end{tabular}
\end{figure}

\paragraph{Algorithm}

Algorithm~\ref{alg:learn} shows our learning algorithm. 
We iterate over the data $\epochs$ times. In each epoch, for each training example $(\instruction^{(i)}, \startstate^{(i)}, \execution^{(i)})$, $i = 1\dots N$, we perform a rollout using our policy to generate an execution (lines~\ref{alg:learn:rollout:start}~-~\ref{alg:learn:rollout:end}). 
The length of the rollout is bound by $\exechorizon$, but may be shorter if the agent selected the $\stopaction$ action. 
At each step $j$, the agent updates the agent context $\ostate_j$ (lines \ref{alg:learn:image}~-~\ref{alg:learn:observe}), samples an action from the policy $\policy$ (line~\ref{alg:learn:sample}), and executes it to generate the new world state $\state_{j+1}$ (line~\ref{alg:learn:execute}). 
The gradient is approximated using the sampled action with the computed reward $\reward^{(i)}(\state_j, \action_j)$ (line~\ref{alg:learn:backpropagate}). 
Following each rollout, we update the parameters $\param$ with the mean of the gradients using \textsc{Adam}~\citep{Kingma:14adam}.

\jle{}{
\paragraph{Discussion} 
Our learning algorithm emphasizes exploration and learns through a simple reward function, which provides positive reward only on task completion, and otherwise gives a small constant negative reward. 
In early stages of training, we observe the agent receives only negative rewards due to the long sequence of dependent actions required to complete the task. 
As a result, the agent learns a policy that always assigns the entire probability mass to the same action irrespective of the observation, and stops exploring. 
In Section~\ref{sec:reshape}, we describe a reward shaping method that uses the complete demonstration trajectory to effectively overcome these issues. }

\section{Reward Shaping}
\label{sec:reshape}

Reward shaping is a method for transforming a reward function by adding a \emph{shaping term} to the problem reward. 
The goal is to generate more informative updates by adding information to the reward. 
We use this method to leverage the training demonstrations, a common form of supervision for training systems that map language to actions. 
Reward shaping allows us to fully use this type of supervision in a reinforcement learning framework, and effectively combine learning from demonstrations and exploration.

Adding an arbitrary shaping term can change the optimality of policies and modify the original problem, for example by making bad policies according to the problem reward optimal according to the shaped function.\footnote{For example, adding a shaping term $F = -R$ will result in a shaped reward that is always 0, and any policy will be trivially optimal with respect to it.} 
\citet{Ng:99rewardshaping} and \citet{Wiewiora:03reshaping} outline potential-based terms that realize sufficient conditions for \emph{safe} shaping.\footnote{For convenience, we briefly overview the theorems of \citet{Ng:99rewardshaping} and \citet{Wiewiora:03reshaping} in Appendix~\ref{sec:app:shaping}.}  
Adding a shaping term is safe if the order of policies according to the shaped reward is identical to the order according to the original problem reward.
While safe shaping only applies to optimizing the total reward, we show empirically the effectiveness of the safe shaping terms we design in a contextual bandit setting.

We introduce two shaping terms. The final shaped reward is a sum of them and the problem reward. 
Similar to the problem reward, we define example-specific shaping terms. 
We  modify the reward function signature as required.

\begin{figure}
	\centering
	\includegraphics[width=0.85\linewidth,clip,trim=2 437 525 12]{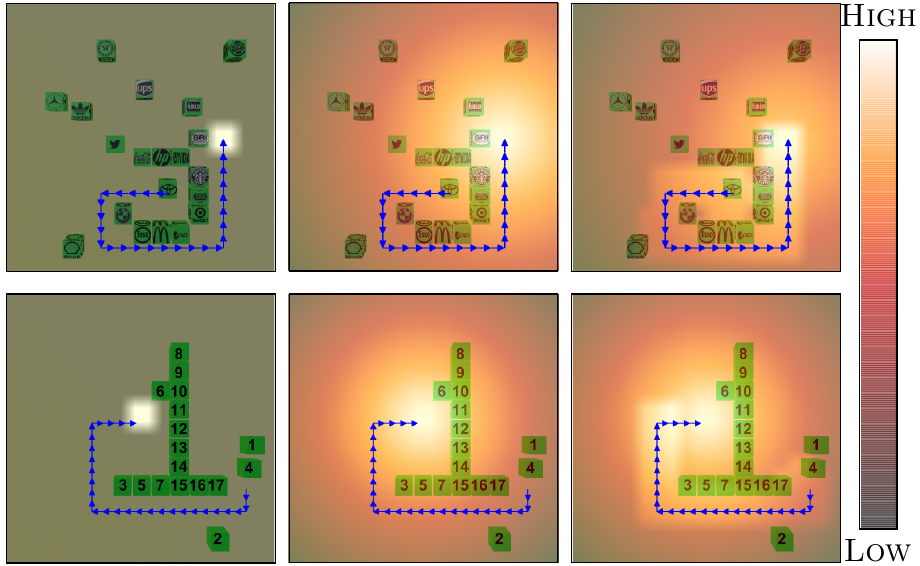}
	\caption{Visualization of the shaping potentials for two tasks. We show   demonstrations (blue arrows), but omit instructions. To visualize the potentials intensity, we assume only the target block can be moved, while rewards and potentials are computed for any block movement. We illustrate the sparse problem reward (left column) as a potential function and consider only its positive component, which is focused on the goal. The middle column adds the distance-based potential. The right adds both potentials. }
	\label{fig:reward}
\end{figure}

\paragraph{Distance-based Shaping ($F_1$)}
The first shaping term measures if the agent moved closer to the goal state. 
We design it to be a safe potential-based term~\cite{Ng:99rewardshaping}:

\begin{small}
\begin{equation}
\nonumber F^{(i)}_1(\state_j, \action_j, \state_{j+1}) =  \phi_1^{(i)}(\state_{j+1})  - \phi_1^{(i)}(\state_j)\;\;.
\end{equation}	
\end{small}

\noindent
The  potential $\phi_1^{(i)}(\state)$ is proportional to the negative distance from the goal state $\goalstate^{(i)}$. 
Formally, \mbox{$\phi_1^{(i)}(\state) = -\eta\|\state-\state_g^{(i)}\|$}, where $\eta$ is a constant scaling factor, and $\|.\|$ is a distance metric. 
In the block world, the distance between two states is the sum of the Euclidean distances between the positions of each block in the two states, and $\eta$ is the inverse of block width.
The middle column in Figure~\ref{fig:reward} visualizes the potential $\phi_1^{(i)}$.

\paragraph{Trajectory-based Shaping ($F_2$)} 

Distance-based shaping may lead the agent to sub-optimal states, for example when an obstacle blocks the direct path to the goal state, and the agent must temporarily increase its distance from the goal to bypass it. 
We incorporate complete trajectories by using a simplification of the shaping term introduced by \citet{Brys:15rl-lfd-shaping}. 
Unlike $F_1$, it requires access to the previous state and action. 
It is based on the look-back advice shaping term of \citet{Wiewiora:03reshaping}, who introduced safe potential-based shaping that considers the previous state and action. 
The second  term is:

\begin{small}
\begin{equation}
\nonumber	F^{(i)}_2(\state_{j-1}, \action_{j-1}, \state_j, \action_j) = \phi_2^{(i)}(\state_j, \action_j) -  \phi_2^{(i)}(\state_{j-1}, \action_{j-1})\;\;.	
\end{equation}	
\end{small}

\noindent
Given $\execution^{(i)} = \langle (\startstate, \action_1), \dots, (\state_m, \action_m)\rangle$, to compute the potential $\phi_2^{(i)}(\state, \action)$, we identify the closest state $\state_j$ in $\execution^{(i)}$ to $\state$. If \mbox{$\eta\|\state_j - \state\| <  1$} and $\action_j = \action$, $\phi_2^{(i)}(\state, \action) = 1.0$, else $\phi_2^{(i)}(\state, \action) = -\rewardpenalty_f$, where $\rewardpenalty_f$ is a penalty parameter. 
We use the same distance computation and parameter $\eta$ as in $F_1$. 
When the agent is in a state close to a demonstration state, this term encourages taking the action taken in the related demonstration state. 
The right column in Figure~\ref{fig:reward} visualizes the effect of the potential $\phi_2^{(i)}$.

\section{Experimental Setup}
\label{sec:exp}

\paragraph{Environment} 
We use the environment of \citet{Bisk:16nl-robots}. 
The original task required predicting the source and target positions for a single block given an instruction.
In contrast, we address the task of moving blocks on the plane to execute instructions given visual input. 
This requires generating the complete sequence of actions needed to complete the instruction. 
The environment contains up to 20 blocks marked with logos or digits. 
Each block can be moved in four directions. Including the $\stopaction$ action, in each step, the agent selects between 81 actions. 
The set of actions is constant and is not limited to the blocks present. The transition function is deterministic.  
The size of each block step is 0.04 of the board size. 
The agent observes the board from above. 
We adopt a relatively challenging setup with a large action space. 
While a simpler setup, for example decomposing the problem to source and target prediction and using a planner, is likely to perform better, we aim to minimize task-specific assumptions and engineering of separate modules. 
However, to better understand the problem, we also report results for the decomposed task with a planner.

\paragraph{Data}
\citet{Bisk:16nl-robots} collected a corpus of instructions paired with start and goal states. 
Figure~\ref{fig:instruction} shows example instructions. 
The original data includes instructions for moving one block or multiple blocks. 
Single-block instructions are relatively similar to navigation instructions and referring expressions.
While they present much of the complexity of natural language understanding and grounding, they rarely display the planning complexity of multi-block instructions, which are beyond the scope of this paper. 
Furthermore, the original data does not include demonstrations. 
While generating demonstrations for moving a single block is straightforward, disambiguating action ordering when multiple blocks are moved is challenging. 
Therefore, we focus on instructions where a single block changes its position between the start and goal states, and restrict demonstration generation to move the changed block. 
The remaining data, and the  complexity it introduces, provide an important direction for future work. 

To create demonstrations, we compute the shortest paths. 
While this process may introduce noise for instructions that specify specific trajectories (e.g., \nlstring{move SRI two steps north and \dots}) rather than only describing the goal state, analysis of the data shows this issue is limited. Out of 100 sampled instructions, 92 describe the goal state rather than the trajectory. 
A secondary source of noise is due to discretization of the state space. As a result, the agent often can not reach the exact target position. 
The demonstrations error illustrates this problem (Table~\ref{tbl:test-results}). 
To provide task completion reward during learning, we relax the state comparison, and consider states to be equal if the sum of block distances is under the size of one block. 

The corpus includes 11,871/1,719/3,177 instructions for training/development/testing. 
Table~\ref{tab:corpus} shows corpus statistic compared to the commonly used SAIL navigation corpus~\cite{MacMahon:06,Chen:11}.
While the SAIL agent only observes its immediate surroundings, overall the blocks domain provides more complex instructions. 
Furthermore, the SAIL environment includes only 400 states, which is insufficient for generalization with vision input. 
We compare to other data sets in Appendix~\ref{sec:app:data}.

\begin{table}[t]
	\footnotesize
	\centering
	\begin{tabular}{|c|c|c|}
	\hline
	& SAIL & Blocks \\
	\hline
	Number of instructions & 3,237 & 16,767 \\
	Mean instruction length & 7.96 & 15.27 \\
	Vocabulary & 563 & 1,426 \\
	Mean trajectory length & 3.12 & 15.4 \\
	\hline
	\end{tabular}
	\caption{Corpus statistics for the block environment we use and the SAIL navigation domain.}
	\label{tab:corpus}
\end{table}

\paragraph{Evaluation}
We evaluate task completion error as the sum of Euclidean distances for each block between its position at the end of the execution and in the gold goal state. We divide distances by block size to normalize for the image size. 
In contrast, \citet{Bisk:16nl-robots} evaluate the selection of the source and target positions independently.

\paragraph{Systems}

We report performance of ablations, the upper bound of following the demonstrations (Demonstrations), and five baselines: 
(a)~$\textsc{Stop}$: the agent immediately stops,
(b)~$\textsc{Random}$: the agent takes random actions, 
(c)~$\textsc{Supervised}$: supervised learning with maximum-likelihood estimate using demonstration state-action pairs, 
(d)~$\textsc{DQN}$: deep Q-learning with both shaping terms~\cite{Mnih:15humanlevelatari}, and
(e)~$\textsc{REINFORCE}$: policy gradient with cumulative episodic reward with both shaping terms~\cite{Sutton:99policygradint}.
Full system details are given in Appendix~\ref{sec:app:baselines}.

\paragraph{Parameters and Initialization}   

Full details are in Appendix~\ref{sec:app:params}.
We consider $\statehorizon=4$ previous images, and horizon length $\exechorizon=40$.
We initialize our model with the $\textsc{Supervised}$ model.

\section{Results}
\label{sec:results}

\begin{table}[t]
\centering
\footnotesize
\begin{tabular}{| p{2.4cm} | c | c |c|c|}
\hline
\multirow{2}{*}{\textbf{Algorithm}} & \multicolumn{2}{c|}{\textbf{Distance Error}} & \multicolumn{2}{c|}{\textbf{Min. Distance}}\\
\cline{2-5}
& \textbf{Mean} & \textbf{Med.}& \textbf{Mean} & \textbf{Med.} \\
\hline
\hline
Demonstrations  & 0.35 & 0.30 & 0.35 & 0.30 \\
\hline
\hline
\multicolumn{5}{|l|}{Baselines} \\
\hline
$\textsc{Stop}$                       &  5.95 & 5.71 & 5.95 & 5.71 \\
$\textsc{Random}$                 & 15.3 & 15.70 & 5.92 & 5.70\\
$\textsc{Supervised}$            & 4.65 & 4.45 &  3.72 &  3.26 \\
$\textsc{REINFORCE}$ & 5.57& 5.29& 4.50 & 4.25 \\
$\textsc{DQN}$                     & 6.04 & 5.78 & 5.63 & 5.49 \\
\hline
\hline
Our Approach & 3.60 & 3.09 & 2.72 &  2.21  \\
\hspace{0.5em}w/o Sup. Init & 3.78 & 3.13 & 2.79 & 2.21 \\
\hspace{0.5em}w/o Prev. Action & 3.95 & 3.44 & 3.20 & 2.56 \\
\hspace{0.5em}w/o  $F_1$  & 4.33 &  3.74 & 3.29 & 2.64 \\
\hspace{0.5em}w/o $F_2$ & 3.74 & 3.11 & 3.13 & 2.49\\
\hspace{0.5em}w/ Distance     & 8.36 & 7.82 & 5.91 & 5.70\\
\hspace{1em}  Reward & & & & \\
\hline
\hline
\multicolumn{5}{|l|}{Ensembles} \\
\hline
$\textsc{Supervised}$ & 4.64 & 4.27 & 3.69 & 3.22 \\
$\textsc{REINFORCE}$ & 5.28 & 5.23 & 4.75 & 4.67 \\
$\textsc{DQN}$ &  5.85 & 5.59 & 5.60 & 5.46 \\
Our Approach & \textbf{3.59} & \textbf{3.03} & \textbf{2.63} & \textbf{2.15}\\
\hline
\end{tabular}
\caption{Mean and median (Med.) development results.}
\label{tbl:dev-results}
\vspace{8pt}
\end{table}

\begin{table}[t]
\centering
\footnotesize
\begin{tabular}{| p{2.4cm} | c | c |c|c|}
\hline
\multirow{2}{*}{\textbf{Algorithm}} & \multicolumn{2}{c|}{\textbf{Distance Error}} & \multicolumn{2}{c|}{\textbf{Min. Distance}}\\
\cline{2-5}
& \textbf{Mean} & \textbf{Med.} & \textbf{Mean} & \textbf{Med.} \\
\hline
\hline
Demonstrations  & 0.37 & 0.31 & 0.37 & 0.31  \\
\hline
\hline
$\textsc{Stop}$                                   & 6.23 & 6.12 & 6.23 & 6.12 \\
$\textsc{Random}$                   & 15.11 & 15.35 & 6.21 & 6.09\\
\hline
\hline
\multicolumn{5}{|l|}{Ensembles} \\
\hline
$\textsc{Supervised}$ & 4.95 & 4.53 & 3.82 & 3.33\\
$\textsc{REINFORCE}$ & 5.69 & 5.57 & 5.11 & 4.99 \\
$\textsc{DQN}$ &  6.15 & 5.97 & 5.86 & 5.77 \\
Our Approach       & \textbf{3.78} & \textbf{3.14} & \textbf{2.83} & \textbf{2.07}\\
\hline
\end{tabular}
\caption{Mean and median (Med.) test results.}
\label{tbl:test-results}
\end{table}

Table~\ref{tbl:dev-results} shows development results. 
We run each experiment three times and report the best result.  
The $\textsc{Random}$ and $\textsc{Stop}$ baselines illustrate the task complexity of the task. 
Our approach, including both shaping terms in a contextual bandit setting, significantly outperforms the other methods. 
$\textsc{Supervised}$ learning demonstrates lower performance. 
A likely explanation is test-time execution errors leading to unfamiliar states with poor later performance~\citep{Kakade:02approximate-rl}, a form of the covariate shift problem.
The low performance of $\textsc{REINFORCE}$ and $\textsc{DQN}$ illustrates the challenge of general reinforcement learning with limited data due to relatively high sample complexity~\cite{Kearns:99sparse-sample-complexity,Krishnamurthy:16PACRL}. 
We also report results using ensembles of the three models.

\begin{figure}[t!]
\begin{center}
\begin{tikzpicture}
 \begin{axis}[
 	width=1\columnwidth,
	    height=0.4\columnwidth,
		font=\scriptsize,
	    xmin=0,xmax=100,
   		ymin=3.5, ymax=5.5,
	    xtick={0,10,20,30,40,50,60,70,80,90,100},
   	    ytick={3.5,4.0,4.5,5.0,5.5},
	    bar width=12pt,
    	xlabel style={yshift=0ex,},
        xlabel=\% Demonstrations,
        ylabel=Mean Error]
		
	\addplot[smooth,mark=*,color=blue] coordinates {
         (0,5.45)
         (12.5,3.9)
         (25,3.76)
    	    (50, 3.69)
         (100,3.60)
     };		

    \end{axis}
\end{tikzpicture}
\end{center}
\caption{Mean distance error as a function of the ratio of training examples that include complete trajectories. The rest of the data includes the goal state only.}
\label{fig:semisup}
\end{figure}
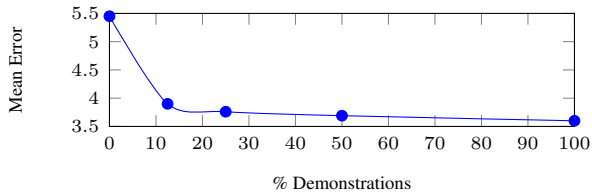

We ablate different parts of our approach. 
Ablations of supervised initialization (our approach w/o sup. init) or the previous action (our approach w/o prev. action) result in increase in error. 
While the contribution of initialization is modest, it provides faster learning. 
On average, after two epochs, we observe an error of $3.94$ with initialization and $6.01$ without. 
We hypothesize that the $F_2$ shaping term, which uses full demonstrations, helps to narrow the gap at the end of learning. Without supervised initialization and $F_2$, the error increases to $5.45$ (the 0\% point in Figure~\ref{fig:semisup}). 
We observe the contribution of each shaping term and their combination. 
To study the benefit of potential-based shaping, we experiment with a negative distance-to-goal reward.
This reward replaces the problem reward and encourages getting closer to the goal (our approach w/distance reward).  
With this reward, learning fails to converge, leading to a relatively high error. 

Figure~\ref{fig:semisup} shows our approach with varying amount of supervision. 
We remove demonstrations from both supervised initialization and the $F_2$ shaping term. For example, when only 25\% are available, only 25\% of the data is available for initialization and the $F_2$ term is only present for this part of the data. 
While some demonstrations are necessary for effective learning, we get most of the benefit with only 12.5\%.

Table~\ref{tbl:test-results} provides test results, using the ensembles to decrease the risk of overfitting the development. 
We observe similar trends to development result with our approach outperforming all baselines.
The remaining  gap to the demonstrations upper bound  illustrates the need for future work.

To understand performance better, we measure \emph{minimal distance} (min. distance in Tables \ref{tbl:dev-results} and \ref{tbl:test-results}), the closest the agent got to the goal. 
We observe a strong trend: the agent often gets close to the goal and fails to stop. 
This behavior is also reflected in the number of steps the agent takes. 
While the mean number of steps in development demonstrations is $15.2$, the agent generates on average $28.7$ steps, and $55.2\%$ of the time it takes the maximum number of allowed steps ($40$). 
Testing on the training data shows an average $21.75$ steps and exhausts the number of steps $29.3\%$ of the time. The mean number of steps in training demonstrations is $15.5$. 
This  illustrates the challenge of learning how to be behave at an absorbing state, which is observed relatively rarely during training. 
This behavior also shows in our video.\footnote{\url{https://github.com/clic-lab/blocks}}

We also evaluate a supervised learning variant that assumes a perfect planner.\footnote{As there is no sequence of decisions, our reinforcement approach is not appropriate for the planner experiment. The architecture details are described in Appendix~\ref{sec:app:baselines}.} This setup is similar to \citet{Bisk:16nl-robots}, except using raw image input. 
It allows us to roughly understand how well the agent generates actions. 
We observe a mean error of $2.78$ on the development set, an improvement of almost two points over supervised learning with our approach. 
This illustrates the complexity of the complete problem.

We conduct a shallow linguistic analysis to understand the agent behavior with regard to differences in the language input. 
As expected, the agent is sensitive to unknown words. 
For instructions without unknown words, the mean development error is $3.49$. 
It increases to $3.97$ for instructions with a single unknown word, and to $4.19$ for two.\footnote{This trend continues, although the number of instructions is too low ($<20$) to be reliable.} 
We also study the agent behavior when observing new phrases composed of known words by looking at  instructions with new n-grams and no unknown words. 
We observe no significant correlation between performance and new bi-grams and tri-grams. 
We also see no meaningful correlation between instruction length and performance. 
Although counterintuitive given the linguistic complexities of longer instructions, it aligns with results in machine translation~\cite{Luong:15nmtattention}.

\section{Conclusions}
\label{sec:conclusions}

We study the problem of learning to execute instructions in a situated environment given only raw visual observations. 
Supervised approaches do not explore adequately to handle test time errors, and reinforcement learning approaches require a large number of samples for good convergence.
Our solution provides an effective combination of both approaches: reward shaping to create relatively stable optimization in a contextual bandit setting, which takes advantage of a signal similar to supervised learning, with a reinforcement basis that admits substantial exploration and easy avenues for smart initialization. 
This combination is designed for a few-samples regime, as we address. 
When the number of samples is unbounded, the drawbacks observed in this scenario for optimizing longer term reward do not hold.

\section*{Acknowledgments}

This research was supported by a Google Faculty Award, an Amazon Web Services Research Grant, and a Schmidt Sciences Research Award. 
We thank Alane Suhr, Luke Zettlemoyer, and the anonymous reviewers for their helpful feedback, and Claudia Yan for technical help. 
We also thank the Cornell NLP group and the Microsoft Research Machine Learning NYC group for their support and insightful comments.

\balance
\bibliography{main}
\bibliographystyle{emnlp_natbib}

\clearpage

\nobalance

\appendix

\section{Reward Shaping Theorems}
\label{sec:app:shaping}

In Section~\ref{sec:reshape}, we introduce two reward shaping terms. 
We follow the safe-shaping theorems of \citet{Ng:99rewardshaping} and \citet{Wiewiora:03reshaping}. 
The theorems outline potential-based terms that realize sufficient conditions for \emph{safe} shaping. 
Applying safe terms guarantees the order of policies according to the original problem reward does not change. 
While the theory only applies when optimizing the total reward, we show empirically the effectiveness of the safe shaping terms in a contextual bandit setting.
For convenience, we provide the definitions of potential-based shaping terms and the theorems introduced by \citet{Ng:99rewardshaping} and \citet{Wiewiora:03reshaping} using our notation. 
We refer the reader to the original papers for the full details and proofs. 

The distance-based shaping term $F_1$ is defined based on the theorem of \citet{Ng:99rewardshaping}:\\[-10pt]

\noindent
\fbox{
\begin{minipage}{0.95\linewidth}
\begin{footnotesize}
\begin{definition*}
A shaping term $\shaping : \allstate \times \allaction \times \allstate \rightarrow \reals$ is potential-based if there exists a function $\phi: \allstate \rightarrow \reals$ such that, at time $j$, $F(\state_j, \action_j, \state_{j+1}) =  \gamma\phi(\state_{j+1}) - \phi(\state_{j})$, $\forall \state_j, \state_{j+1} \in \allstate$ and $\action_j \in \allaction$, where $\gamma \in [0,1]$ is a future reward discounting factor. The function $\phi$ is the potential function of the shaping term $F$. 
\end{definition*}
\begin{theorem*}
Given a reward function $\reward(\state_j, \action_j)$, if the shaping term is potential-based, the shaped reward $\reward_\shaping(\state_j, \action_j, \state_{j+1}) = R(\state_j, \action_j) + F(\state_j, \action_j, \state_{j+1})$ does not modify the total order of policies. 
\end{theorem*}
\end{footnotesize}
\end{minipage}
}

\noindent
In the definition of $F_1$, we set the discounting term $\gamma$ to 1.0 and omit it. 

The trajectory-based shaping term $F_2$ follows the shaping term introduced by \citet{Brys:15rl-lfd-shaping}. 
To define it, we use the look-back advice shaping term of \citet{Wiewiora:03reshaping}, who extended the potential-based term of \citet{Ng:99rewardshaping} for terms that consider the previous state and action:\\[-10pt]

\noindent
\fbox{
\begin{minipage}{0.95\linewidth}
\begin{footnotesize}
\begin{definition*}
A shaping term $\shaping: \allstate \times \allaction \times \allstate \times \allaction \rightarrow \reals$ is potential-based if there exists a function $\phi: \allstate \times \allaction \rightarrow \reals$ such that, at time $j$, $F(\state_{j-1}, \action_{j-1}, \state_j, \action_j) =  \gamma\phi(\state_j, \action_j) -  \phi(\state_{j-1}, \action_{j-1})$, $\forall \state_j, \state_{j-1} \in \allstate$ and $\action_j, \action_{j-1} \in \allaction$, where $\gamma \in [0,1]$ is a future reward discounting factor. The function $\phi$ is the potential function of the shaping term $F$. 
\end{definition*}
\begin{theorem*}
Given a reward function $\reward(\state_j, \action_j)$, if the shaping term is potential-based, the shaped reward $\reward_\shaping(\state_{j-1}, \action_{j-1}, \state_j, \action_j) = R(\state_j, \action_j) + F(\state_{j-1}, \action_{j-1}, \state_j, \action_j)$ does not modify the total order of policies. 
\end{theorem*}
\end{footnotesize}
\end{minipage}
}

\noindent
In the definition of $F_2$ as well, we set the discounting term $\gamma$ to 1.0 and omit it.

\section{Evaluation Systems}
\label{sec:app:baselines}

We implement multiple systems for evaluation. 

\paragraph{$\textsc{Stop}$} 

The agent performs the $\stopaction$ action immediately at the beginning of execution.

\paragraph{$\textsc{Random}$} 

The agent samples actions uniformly until $\stopaction$ is sampled or $\exechorizon$ actions were sampled, where $\exechorizon$ is the execution horizon.

\paragraph{$\textsc{Supervised}$} 

Given the training data with $N$  instruction-state-execution triplets, we generate training data of instruction-state-action triplets and optimize the log-likelihood of the data. 
Formally, we optimize the objective:

\begin{small}
\begin{equation}
\nonumber	J = \frac{1}{N}\sum_{i=1}^N \sum_{j=1}^{m^{(i)}} \log\policy(\ostate_j^{(i)}, \action_j^{(i)})\;\;,
\end{equation}
\end{small}

\noindent
where $m^{(i)}$ is the length of the execution $\execution^{(i)}$, $\ostate_j^{(i)}$ is the agent context at step $j$ in sample $i$, and $\action_j^{(i)}$ is the demonstration action of step $j$ in demonstration execution $\execution^{(i)}$. 
Agent contexts are generated with the annotated previous actions (i.e., to generate previous images and the previous action). 
We use minibatch gradient descent with \textsc{Adam} updates~\cite{Kingma:14adam}.

\paragraph{$\textsc{DQN}$} 

We use deep Q-learning~\cite{Mnih:15humanlevelatari} to train a Q-network. 
We use the architecture described in Section~\ref{sec:model}, except replacing the task specific part with a single 81-dimension layer. In contrast to our probabilistic model, we do not decompose block and direction selection. 
We use the shaped reward function, including both $F_1$ and $F_2$. 
We use a replay memory of size 2,000 and an $\epsilon$-greedy behavior policy to generate rollouts. We attenuate the value of $\epsilon$ from 1 to 0.1 in 100,000 steps and use prioritized sweeping for sampling. We also use a target network that is synchronized after every epoch.

\paragraph{$\textsc{REINFORCE}$}

We use the $\textsc{REINFORCE}$ algorithm \citep{Sutton:99policygradint}  to train our agent. $\textsc{REINFORCE}$ performs policy gradient learning with total reward accumulated over the roll-out as opposed to using immediate rewards as in our main approach. $\textsc{REINFORCE}$ samples the total reward using monte-carlo sampling by performing a roll-out. We use the shaped reward function,  including both $F_1$ and $F_2$ terms.
Similar to our approach, we initialize with a $\textsc{Supervised}$ model and regularize the objective with the entropy of the policy. We do not use a reward baseline.

\paragraph{\textsc{Supervised} with Oracle Planner}

We use a variant of our model assuming a perfect planner. The model predicts the block to move and its target position as a pair of coordinates. 
We modify the architecture in Section~\ref{sec:model} to predict the block to move and its target position as a pair of coordinates. 
This model assumes that the sequence of actions is inferred from the predicted target position using an oracle planner.
We train  using supervised learning by maximizing the likelihood of the  block being moved and minimizing the squared distance between the predicted target position and the annotated target position.

\section{Parameters and Initialization}
\label{sec:app:params}

\subsection{Architecture Parameters}

We use an RGB image of 120x120 pixels, and a convolutional neural network ($\conv$) with 4 layers. 
The first two layers apply 32 $8\times8$ filters with a stride of 4, the third applies 32 $4\times4$ filters with a stride of 2. 
The last layer performs an affine transformation to create a 200-dimension vector. 
We linearly scale all images to have zero mean and unit norm.
We use a single layer RNN with 150-dimensional word embeddings and 250 LSTM units. 
The dimension of the action embedding $\embed_a$ is $56$, including 32 for embedding the block and 24 for embedding the directions. $\mathbf{W}^{(1)}$ is a $506\times 120$ matrix and $\mathbf{b}^{(1)}$ is a 120-dimension vector. $\mathbf{W}^{(D)}$ is $120\times 20$ for 20 blocks, and $\mathbf{W}^{(B)}$ is $120\times 5$ for the four directions (north, south, east, west) and the $\stopaction$ action.
We consider $\statehorizon=4$ previous images, and use horizon length $\exechorizon=40$.

\subsection{Initialization}

Embedding matrices are initialized with a zero-mean unit-variance Gaussian distribution.
All biases are initialized to $\mathbf{0}$.  We use a zero-mean truncated normal distribution to initialize the CNN filters (0.005 variance) and CNN weights matrices (0.004 variance). All other weight matrices are initialized with a normal distribution (mean=$0.0$, standard deviation=$0.01$). 
The matrices used in the word embedding function $\embed$ are initialized with a zero-mean normal distribution with standard deviation of 1.0. Action embedding matrices, which are used for $\embed_a$, are initialized with a zero-mean normal distribution with 0.001 standard deviation.
We initialize policy gradient learning, including our approach, with parameters estimated using supervised learning for two epochs, except the direction parameters $\mathbf{W}^{(D)}$ and $\mathbf{b}^{(D)}$, which we learn from scratch. 
We found this initialization method to provide a good balance between strong initialization and not biasing the learning too much, which can result in limited exploration. 

\subsection{Learning Parameters}

We use the distance error on a small validation set as stopping criteria. 
After each epoch, we save the model, and select the final model based on development set performance. 
While this method overfits the development set, we found it more reliable then using the small validation set alone. Our relatively modest performance degradation on the held-out set illustrates that our models generalize well. 
We set the reward and shaping penalties $\rewardpenalty=\rewardpenalty_f=0.02$. 
The entropy regularization coefficient is $\entropycoef=0.1$.
The learning rate is $\mu=0.001$ for supervised learning and $\mu=0.00025$ for policy gradient.  We clip the gradient at a norm of $5.0$. All learning algorithms use a mini-batch of size 32 during training.

\section{Dataset Comparisons}
\label{sec:app:data}

\begin{table*}
	\centering
	\footnotesize
	\begin{tabular}{|p{2cm}|c|c|c|c|c|c|}
		\hline
		Name & \# Samples & Vocabulary & Mean Instruction   & \# Actions & Mean Trajectory  & Partially   \\
		& & Size & Length & &   Length  & Observed  \\
		\hline
		Blocks & 16,767& 1,426 & 15.27 & 81 & 15.4 & No  \\
		\dline
		SAIL & 3,237 & 563 & 7.96 &3 & 3.12 & Yes  \\
		\dline
		Matuszek & 217 & 39 & 6.65 & 3 & N/A & No  \\
		\dline
		Misra & 469 & 775 & 48.7 & $>100$ & 21.5 & No   \\
		\hline
	\end{tabular}
	\caption{Comparison of several related natural language instructions corpora.}
	\label{tab:corpora}
\end{table*}

We briefly review instruction following datasets in Table~\ref{tab:corpora}, including:  Blocks~\cite{Bisk:16nl-robots}, SAIL~\cite{MacMahon:06,Chen:11}, Matuszek~\cite{Matuszek:12b}, and Misra~\cite{Misra:15highlevel}.
Overall, Blocks provides the largest training set and a relatively complex environment with well over $2.43^{18}$ possible states.\footnote{We compute this loose lower bound on the number of states in the block world as $20! = 2.43^{18}$ (the number of block permutations). This is a very loose lower bound.} 
The most similar dataset is SAIL, which provides only partial observability of the environment (i.e., the agent observes what is around it only). However, SAIL is less complex on other dimensions related to the instructions, trajectories, and action space. 
In addition, while Blocks has a large number of possible states, SAIL includes only 400 states. 
The small number of states makes it difficult to learn vision models that generalize well. 
Misra~\cite{Misra:15highlevel} provides a parameterized action space (e.g., $\rm{grasp}(\rm{cup})$), which leads to a large number of potential actions. However, the corpus is relatively small.

\section{Common Questions}
\label{sec:app:faq}

This is a list of potential questions following various decisions that we made. While we ablated and discussed all the crucial decisions in the paper, we decided to include this appendix to provide as much information as possible.

\paragraph{Is it possible to manually engineer a competitive reward function without shaping?}

Shaping is a principled approach to add information to a problem reward with relatively intuitive potential functions. Our experiments demonstrate its effectiveness. 
Investing engineering effort in designing a  reward function specifically designed to the task is a potential alternative approach.

\paragraph{Are you using beam search? Why not?}

While using beam search can probably increase our performance, we chose to avoid it. 
We are motivated by robotic scenarios, where implementing  beam search is a challenging task and often not possible. 
We distinguish between beam search and back-tracking. 
Beam search is also incompatible with common assumptions of reinforcement learning, although it is often used during test with reinforcement learning systems. 

\paragraph{Why are you using the mean of the LSTM hidden states instead of just the final state?} 

We empirically tested both options. Using the mean worked better. This was also observed by \citet{Narasimhan:15text-games}. Understanding in which scenarios one technique is better than the other is an important question for future work. 

\paragraph{Can you provide more details about initialization?}

Please see Appendix~\ref{sec:app:params}.

\paragraph{Does the agent in the block world learn to move obstacles and other blocks?}

While the agent can move any block at any step, in practice, it rarely happens. The agent prefers to move blocks around obstacles rather than moving other blocks and moving them back into place afterwards. 
This behavior is learned from the data and shows even when we use only very limited amount of demonstrations. 
We hypothesize that in other tasks the agent is likely to learn that moving obstacles is advantageous, for example when demonstrations include moving obstacles. 

\paragraph{Does the agent explicitly mark where it is in the instruction?}

We estimate that over 90\% of the instructions describe the target position.  
Therefore, it is often not clear how much of the instruction was completed during the execution. 
The agent does not have an explicit mechanism to mark portions of the instruction that are complete. 
We briefly experimented with attention, but found that empirically it does not help in our domain. 
Designing an architecture to allows such considerations is an important direction for future work. 

\paragraph{Does the agent know which blocks are present?}

Not all blocks are included in each task. The agent must infer which blocks are present from the image and instruction. The set of possible actions, which includes moving all possible blocks, does not change between tasks. If the agent chooses to move a block that is not present, the world state does not change. 

\paragraph{Did you experiment with executing sequences of instruction? The \citet{Bisk:16nl-robots} includes such instructions, right?}

The majority of existing corpora, including SAIL~\cite{Chen:11,Artzi:13,Mei:16neuralnavi}, provide segmented sequences of instructions. Existing approaches take advantage of this segmentation during training. For example, \citet{Chen:11}, \citet{Artzi:13}, and \citet{Mei:16neuralnavi} all train on segmented data and test on sequences of instructions by doing inference on one sentence at a time. We are also able to do this. Similar to these approaches, we will likely suffer from cascading errors. 
The multi-instruction paragraphs in the \citet{Bisk:16nl-robots} data are an open problem and present new challenges beyond just  instruction length. For example, they often merge multiple block placements in one instruction (e.g, \nlstring{put the SRI, HP, and Dell blocks in a row}). Since the original corpus does not provide trajectories and our automatic generation procedure is not able to resolve which block to move first, we do not have demonstrations for this data. 
The instructions also present a significantly more complex task. This is an important direction for future work, which illustrates the complexity and potential of the domain.

\paragraph{Potential-based shaping was proven to be safe when maximizing the total expected reward. Does this apply for the contextual bandit setting, where you maximize the immediate reward?}

The safe shaping theorems (Appendix~\ref{sec:app:shaping}) do not hold in our contextual bandit setting. We show empirically that shaping works in practice. However, how and if it changes the order of policies is an open question. 

\paragraph{How long does it take to train? How many frames the agent observes?}

The agent observes about 2.5 million frames. It takes 16 hours using 50\% capacity of an Nvidia Pascal Titan X GPU to train using our approach. DQN takes more than twice the time for the same number of epochs. Supervised learning  takes about 9 hours to converge. We also trained DQN for around four days, but did not observe improvement. 

\paragraph{Did you consider initializing DQN with supervised learning?}

Initializing DQN with the probabilistic supervised model is challenging. Since DQN is not probabilistic it is not clear what this initialization means. Smart initialization of DQN is an important problem for future work.

\end{document}